# LoTR: Low Tensor Rank Weight Adaptation

**Daniel Bershatsky** [1] **Daria Cherniuk** [1] **Talgat Daulbaev** [1] **Aleksandr Mikhalev** [1] **Ivan Oseledets** [2 1]

## Abstract

In this paper we generalize and extend an idea of low-rank adaptation (LoRA) of large language models (LLMs) based on Transformer architecture. Widely used LoRA-like methods of fine-tuning LLMs are based on matrix factorization of gradient update. We introduce LoTR, a novel approach for parameter-efficient fine-tuning of LLMs which represents a gradient update to parameters in a form of tensor decomposition. Low-rank adapter for each layer is constructed as a product of three matrices, and tensor structure arises from sharing left and right multipliers of this product among layers. Simultaneous compression of a sequence of layers with low-rank tensor representation allows LoTR to archive even better parameter efficiency then LoRA especially for deep models. Moreover, the core tensor does not depend on original weight dimension and can be made arbitrary small, which allows for extremely cheap and fast downstream fine-tuning.

## 1. Introduction

Pre-trained Transformer-based large language models (LLMs) demonstrates excellent performance in language modeling and other downstream task like sentinel analysis, summarization, part of speech tagging, and so on. Well-established paradigm of solving practical problems in natural language domain is fine-tuning of pre-trained LLM to target domain data. As number of parameters of LLMs grows usual fine-tuning approach becomes more and more costly in every respect.

Recent works (Guo et al., 2021; 2021; He et al., 2022; Houlsby et al., 2019; Hu et al., 2021) study the problem of parameter-effective fine-tuning which requires to train as less as possible adjustable parameters while attaining the best possible performance on a task. The common idea is to consider a set of newly inserted or a subset of existing parameters of language model which should be adjustable to training data (see Section 4).

In this paper we are developing a line of work based on introducing a low-rank correction to weights. In particular, authors in (Hu et al., 2021) proposed fine-tuning of a model with reparametrized weight matrices so that correction term have low-rank structure. Specifically, original weight matrix $W$ is kept frozen but trainable low-rank term $BA^\top$ is summed up with original weight matrix $W$, that is,

$$W \to W + \alpha BA^\top,$$

where matrices $A \in \mathbb{R}^{n_\text{out} \times r}$ and $B \in \mathbb{R}^{n_\text{in} \times r}$ have a small number of columns (i.e. small rank). The number of trainable parameters depends linearly on the input and output size, $O(nr)$, rather than quadratically, making fine-tuning computationally more appealing.

We propose a generalization, called LoTR[1] (Low Tensor Rank adaptation), and an extension of a low-rank adaptation training procedure LoRA (Hu et al., 2021) in order to achieve even better compression rate and parameter efficiency. The core idea is a consideration not a sole single layer but a series of layers in Transformer across Transformer-block to make it possible to represent them as a low-rank tensor factorization rather than matrix factorization.

With our approach to fine-tuning, a number of trainable parameters has a significantly better bound in comparison with LoRA. Namely, LoRA applied to a single layer in $L$ transformer-blocks should to update $O(nrL)$ parameters but LoTR trains only $O(Lr^2 + nr)$ parameters. This means that LoTR exhibits better parameter efficacy than LoRA as a number of transformer-blocks (depth of transformer) increases. It is worth noting that contemporary architectures often consist of tens or even hundreds of Transformer-blocks in their typical depth what makes LoTR event more appealing.

In this study, we introduce LoTR which generalizes LoRA along Transformer-blocks dimension, has better compression rate, and suggests a hyperparameter $r$ (rank of factorization) which allows to change number of trainable parameters more smoothly (see Section 2). We empirically demonstrate that LoTR performs competitively with LoRA and achieve the similar performance on downstream tasks (see Section 3). Finally, we describe our ablation studies and give an intuition on hyperparameter setting.

[1]AI Center, Skoltech, Moscow, Russia [2]AIRI, Moscow, Russia. Correspondence to: Daniel Bershatsky <d.bershatsky2@skoltech.ru>.

[1]Our implementation is available at github.com/skolai/lotr.

## 2. Method

As mention earlier in Section 1, LoTR extends and generalizes LoRA (Hu et al., 2021) in multiple ways. In this section we will introduce LoTR after a short discussion of LoRA and low-rank tensor factorization like Tucker decomposition (Tucker, 1966) or tensor train decomposition (Oseledets, 2011).

### 2.1. Tensor Decompositions

Similar to matrix decomposition there are plenty of tensor decompositions which are well-studied algebraic objects with remarkable practical properties. In this section we introduce Tucker and Tensor Train decompositions which we later use in low-rank tensor adaptation of neural networks.

#### 2.1.1. Tucker Decomposition

Tucker decomposition (Tucker, 1966) can be viewed as a generalization of SVD or PCA for the case of three dimensions. If tensor $X \in \mathbb{R}^{n\times m\times l}$ can be expressed in a form

$$X = G \times_1 A \times_2 B \times_3 C, \tag{1}$$

where tensor $G \in \mathbb{R}^{r_1\times r_2\times r_3}$ is called a core, and $A \in \mathbb{R}^{r_1\times n}$, $B \in \mathbb{R}^{r_2\times m}$, $C \in \mathbb{R}^{r_3\times l}$ are factor matrices, then we say that tensor $X$ is represented as Tucker decomposition with core $G$ and factors $A$, $B$, and $C$.

This definition uses mode-$k$ (matrix) multiplication $\times_k$ which can be written for $d$-tensor $A \in \mathbb{R}^{n_1\times n_2\times \dots\times n_d}$, matrix $B \in \mathbb{R}^{n_k\times l}$, and $k \le d$ as follows,

$$A \times_k B = \sum_{i_k=j_1} A_{i_1 i_2 \dots i_k \dots i_d} B_{j_1 j_2}.$$

Result of mode-$k$ multiplication is certainly a $d$-tensor of the same dimension $d$ and shape $n_1 \times n_2 \times \dots \times n_{k-1} \times l \times n_{k+1} \times \dots \times n_N$. In index notation expression (1) can be written explicitly as follows

$$X_{ijk} = \sum_{p,q,r} G_{pqr} A_{pi} B_{qj} C_{rk}. \tag{2}$$

Decomposition (1) is also known under Tucker3 alias. In Section 2.3 we will need a degenerate case of Tucker decomposition which is called Tucker2 decomposition. It is a special case for trivial factor $C$ when it equals to an identity matrix $C = I$. Thus, equation (1) can be simplified and expressed as

$$\begin{aligned} X &= G \times_1 A \times_2 B \times_3 I \\ &= G \times_1 A \times_2 B, \end{aligned} \tag{3}$$

where tensor $G$ still has 3 modes. One can write down Tucker1 decomposition in the same manner if the second factor matrix $B$ is trivial $B = I$ as well.

Numbers $r = (r_1, r_2, r_3)$ are called $n$-rank (or just rank) of tensor $X$. Tucker decomposition represents a tensor of $O(n^3)$ elements with $O(r^3 + rn)$ elements what is significantly compact in case of $r \ll n$.

#### 2.1.2. Tensor Train Decomposition

Tensor train decomposition of a n-tensor $X$ was introduced in (Oseledets, 2011). Tucker decomposition can be viewed as a specific 3-dimensional case of tensor train decomposition.We say that tensor $X$ can be represented in tensor train form (or TT for brevity) if every element of tensor $X$ can be expressed as a multiplication of cores $G_k$, $k \in \{1, 2, ..., n\}$

$$\begin{aligned} X_{i_1 i_2 \dots i_n} &= \prod_{k=1}^{n} G_k(i_k) \\ &= G_1(i_1) G_2(i_2) \dots G_n(i_n). \end{aligned} \tag{4}$$

Cores $G_k$ are tensors with 3 modes and $G_k \in \mathbb{R}^{r_{k-1}\times r_k\times r_{k+1}}$. Numbers $r_{0,n+1}$ are assumed to be trivial $r_0 = r_{n+1} = 1$, and an ordered tuple $r = (r_1, r_2, ..., r_n)$ is called a rank of TT decomposition. One can see that if tensor $X$ admits TT representation then only $O(dnr^2)$ elements is sufficient to store a tensor of $O(n^d)$ elements in total. In other words, a tensor in TT form can be represented with exponentially smaller number of elements.

### 2.2. Low-Rank Adaptation (LoRA)

LoRA was introduced in (Hu et al., 2021) and suggests another method for parameter-efficient fine-tuning. It represents an update to a particular weight matrix (kernel) in additive form where the first term $W$ is frozen original weights and the second one is low-rank correction $\delta W = BA^\top$. In other words, a linear layer with LoRA adapter acts on input batch $X$ of $N$ samples as follows

$$\begin{aligned} Y &= X(W^\top + \alpha AB^\top) + \mathbb{1}_N b^\top \\ &= \mathrm{Aff}(X) + \alpha XAB^\top, \end{aligned} \tag{5}$$

where $\mathbb{1}_N$ is a vector of $N$ ones. Matrices $A \in \mathbb{R}^{n_\text{in}\times r}$ and $B \in \mathbb{R}^{n_\text{out}\times r}$ are of rank $r$. Moreover, rank $r$ is small enough in comparison to a feature dimension $r \ll n$. In this case, it requires to fit only $2nr$ parameters instead of $n^2$ in conventional full fine-tuning setup. Hyperparameter $\alpha$ is used to balance correction activation and gradients through it with respect to original affine part of fully-connected layer.

The main advantage of LoRA is that it reduces significantly a number of trainable parameters, optimizer state and weight storage while preserving performance of adapted model on the level of fully fine-tuned model. Another interesting property of LoRA in context of building LLM applications is that LoRA allows to reuse efficiently the same base model and weight for different domain specific models which are orchestrated by separate LLM controller (Shen et al., 2023).

In the original work LoRA was applied to weight matrices $W_{q,k,v,o}$ in attention. We will stick to this choice of layers to which an adapter is applied.

### 2.3. Low Tensor Rank Adaptation (LoTR)

As low-rank matrix decomposition – the core piece of LoRA – allows to substantially decrease a number of trainable parameters as low rank tensor decompositions enables LoTR to reduce that number even further. Namely, we consider weight matrices of interest not individually but collectively. This makes it possible to apply a tensor decomposition to correction terms, share some factors among linear layers, and eventually adjust fewer parameters.

Each model of Transformer family has regular structure: there is an architectural block which is repeated $N_{\text{depth}}$ times (usually tens times) and constitutes a body of a Transformer. So the weight matrices of interest $W_{q,k,v,o}$ can be enumerated by an index of a block $s$, $1 \le s \le N_{\text{depth}}$. This defines a whole family of weight matrices $\mathcal{W} = \{W_\alpha\}$. Bearing in mind that different matrices have different sense (e.g. key matrix $W_k$ and output projection $W_o$), it is natural to limit entire family $\mathcal{W}$ to a matrices of a specific kind. For example $\mathcal{W}_k$ stands for a set of key matrices in all blocks. However, the choice of a family of matrices is not restricted at all. The only condition that a set of matrix has a fixed order of elements that is basically consistency condition and necessary for applying tensor methods further.

Let $\mathcal{W}$ be a family of matrices in a Transformer model and consider a set of corrections $\delta\mathcal{W} = \{\delta W_\alpha\}$ for each $W_\alpha \in \mathcal{W}$ as it is done in LoRA but without low-rank constraint (see Section 2.2). Since all matrices $\delta\mathcal{W}$ (and $\mathcal{W}$ as well) have the same shape, an entire family of matrices $\delta\mathcal{W}$ can be viewed as 3-tensor such that

$$(\delta W_\alpha)_{ij} = \delta \mathcal{W}_{ij\alpha},$$

where index $\alpha$ enumerates matrices in the family and $i$, $j$ are row and column indices of an element of a matrix $\delta W_\alpha$.

Then one can eventually apply low-rank constraint and consider Tucker2 decomposition of correction tensor $\delta\mathcal{W}$

*Table 1.* Summary table of scaling laws for different parameter-efficient fine-tuning methods. Depth of a model is $L$, number of hidden units is $d$, total number of parameter of backbone model is $N$. Parameters $r$ and $m$ are specific to a method. While asymptotics are written for encoder-only model like RoBERTa$_{\text{base}}$, a model architecture does not affect them drastically.

| Method | Num. Params |
|---|---|
| Full fine-tuning | $L(8d^2 + 11d)$ |
| BitFit (Ben-Zaken et al., 2022) | $LN_{\text{biases}}d$ |
| Adapter (Houlsby et al., 2019) | $2L(2dm + d + m)$ |
| DiffPrunning (Guo et al., 2021) | $2N$ |
| (IA)³ (Liu et al., 2022) | $8Ld$ |
| LoRA (Hu et al., 2021) | $4Ldr$ |
| LoTR (ours) | $2Lr^2 + 2dr$ |

$$\delta\mathcal{W} = G \times_1 A \times_2 B.$$

In index notation the same expression can be written as follows

$$\delta \mathcal{W}_{ijs} = \sum_{p,q} G_{pqs} A_{pi} B_{qj}. \tag{6}$$

The third index $s$ still enumerates a specific correction $\delta W_s$ to a specific weight matrix $W_s$. Since index $s$ does not take part in any summation in both sides of expression, we can rewrite (6) as

$$\delta W_s = A^\top G_s B, \quad \forall s.$$

The right hand side of the previous expression resembles singular value decomposition (SVD) with the different that matrix $G_s$ is arbitrary and therefore not necessary diagonal.

This motivates an introducing a LoTR (Low Tensor Rank) adaptation technique.

**Definition 2.1.** Given a model $M$ and set of its equally shaped weight matrices $\mathcal{W} = \{W_\alpha\}$, rank $r$ LoTR-adaptation of $M$ is a 3-tensor of corrections $\delta\mathcal{W}$ such that the $s$-th linear layer acts on input batch $X \in \mathbb{R}^{N \times d}$ as

$$Y = \text{Aff}(X) + \alpha X A G_s B^\top, \quad \alpha \in \mathbb{R},$$

where matrices $A \in \mathbb{R}^{d \times r}$ and $B \in \mathbb{R}^{d \times r}$ are shared among all layers in training time while square matrix $S_s \in \mathbb{R}^{r \times r}$ is specific to the $s$-th layer. Number of adjustable parameters equals to $|\mathcal{W}|r^2 + 2rd$.

The number of matrices in the set $\mathcal{W}$ is typically a multiple of model depth $L$ since parameter-efficient training approaches are often applied to matrices of the same kind in each Transformer-block. Following LoRA, we apply LoTR to all matrices $W_q$ and $W_v$, yielding $|\mathcal{W}| = 2L$.

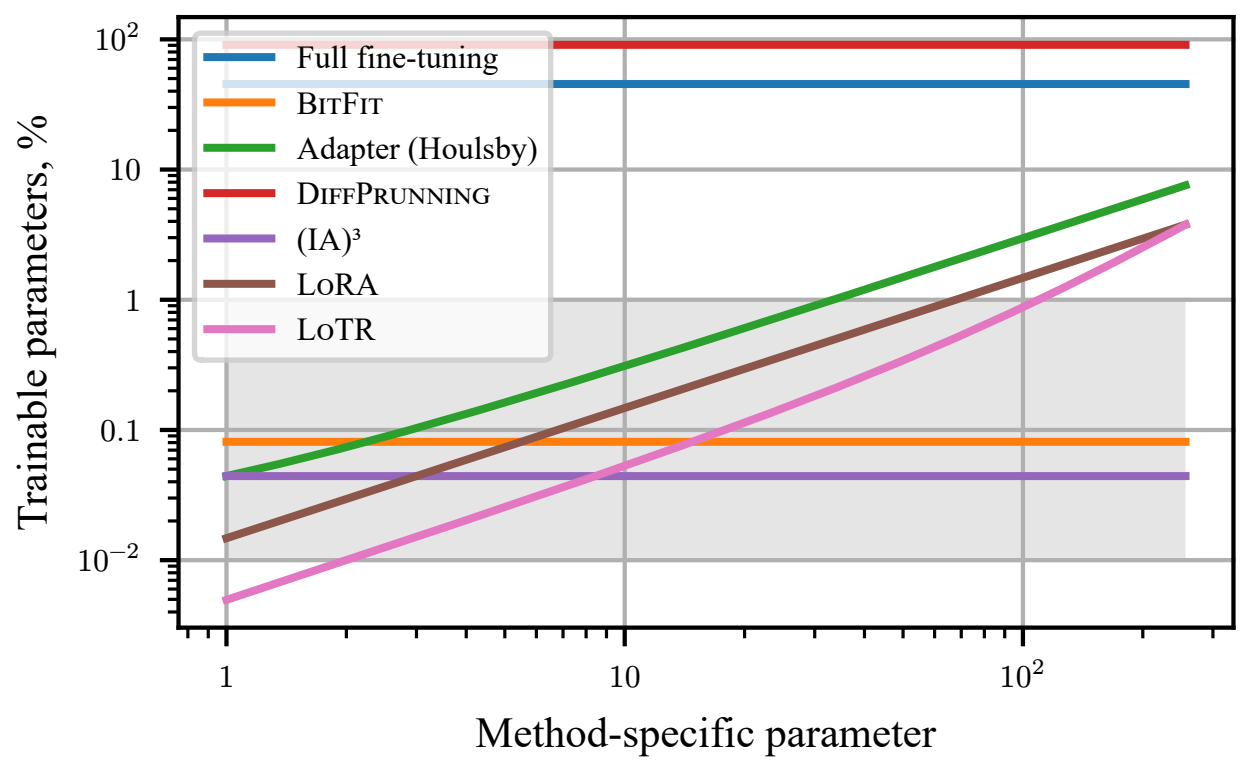

*Figure 1.* Scaling of a number of trainable parameters for various methods applied to RoBERTA$_{\text{base}}$ with reference to a method-specific parameter. Shaded region corresponds to a small number of auxiliary parameters, a setting of practical interest.

Thus the number of updatable parameters in LoTR is $O(Lr^2 + dr)$ which is remarkably different from other methods (see Table 1). Depth $L$ and embedding size $d$ are decoupled due to joint treatment of the entire collection of layers, while a leading term of the most concurrent methods is $O(Ldp)$, where $p$ is a method-specific parameter, e.g. bottleneck $m$ or matrix rank $r$. LoTR demonstrates parameter efficiency when its rank $r$ is below the threshold value $r_{\text{threshold}} \propto \sqrt{dp}$. This threshold is generally high enough to be practical. For example, a realistic value of LoRA rank is $r_{\text{LoRA}} = 8$ that gives a threshold $r_{\text{threshold}} \propto \sqrt{2 \cdot 1024 \cdot 8} = 128$ assuming the use of adapters in RoBERTA$_{\text{large}}$ (see Appendix B). Section 3 reveals that this threshold comfortably accommodates a wide range of reasonable values for $r$.

#### 2.3.1. Initialization

One may come up with several choices for initialization scheme of matrix factors $A$, $B$, and core tensor $S$. We find that the best initialization fills core tensor $S$ with zeros, and draw elements of $A$ and $B$ from standard normal distribution $\mathcal{N}(0, 1)$. Similar initialization scheme should work well with standard initialization methods e.g. (Glorot and Bengio, 2010; He et al., 2015) substituting $\mathcal{N}(0, 1)$.

All initialization approaches for a low-rank adapter are restricted with a condition that an adapter should produce zero activation immediately at initialization. It is worth to note that gradients through an adapter and gradients over adapter weights are not necessarily equal to zero, although the activations are zeros. We consider several initialization schemes and study them empirically in Section 3.2.2.

To specify an initialization one needs (i) to pick a distribution used to draw factor elements from, and (ii) to select a factor that will be zeroed. Since factor matrices and core tensor have different nature, we use different initializers for them. We initialize factor matrices $A$ and $B$ either with zeros or elements sampled from standard normal distribution $\mathcal{N}(0, 1)$, or Tucker2 factors of frozen weights $\mathcal{W}$. Core tensor $S$ is initialized either with zeros or identity matrices such that $S_s = I$, or with core tensor of Tucker2 decomposition obtained frozen weights $\mathcal{W}$.

Our motivation for considering Tucker2 decomposition of $\mathcal{W}$ for initialization is based on study of the relationship between $W$ and $\Delta W$ in LoRA (Hu et al., 2021). The idea is that matrix factors $A$, $B$ initialized from top singular vectors could help training an adapter in early epoches improving convergence. However, it turns out that Tucker2-based initialization performs much worse than factors sampled from standard normal distribution (see Section 3.2.2). This strengthens the hypothesis that optimization happens in directions orthogonal to main singular vectors.

#### 2.3.2. High-Order Generalization

Surely, one should not necessary limit oneself to a 3-tensor corrections $\delta\mathcal{W}$ and can consider higher-order corrections. For example, such generalization might be desirable in a context of mixture of experts (MoE) and models like Mixtral (Jiang et al., 2024). Initially, layers are concatenated along the depth dimension but MoE introduces additional stacking direction. This results in 4 dimensional weight $\mathcal{W}$ and correction $\delta\mathcal{W}$ tensors.

The direct generalization of Tucker decomposition is HOSVD (Kolda and Bader, 2009). However, the curse of dimensionality becomes the issue since core $n$-tensor $G$ contributes the most to the number of trainable parameters. It scales as $O(Lr^n)$, subsequently event a small value of rank $r$ results in significant amount of parameters to train.

Fortunately, there are tensor decompositions which break the curse of dimensionality. One specific example is TT decomposition (see Section 2.1.2). It can be viewed as a generalization of Tucker2 decomposition in the following way. In case of 3 dimensional correction tensor $\delta W$, the first and the last TT-cores $G_1(i_1)$ and $G_3(i_3)$ are equivalent to factor matrices $A$ and $B$ of Tucker2 decomposition. Then the second TT-core $G_2(i_2)$ corresponds to Tucker2-core $G$. In case of $n$ dimensional correction $\delta W$, factor matrices $A$ and $B$ still match the first and the last TT-cores $G_1(i_1)$ and $G_{n(i_n)}$, but tensor core $G$ now corresponds to the inner tensor train $G_2(i_2)G_3(i_3)...G_{n-1}(i_{n-1})$. As it was mentioned in Section 2.1.2, a tensor represented in tensor train format requires only $O(dnr^2)$ elements, i.e. it scales quadratically

*Table 2.* Comparison of different parameter-efficient fine-tuning methods applied to encoder-only TRANSFORMER models on GLUE benchmark. LoTR demonstrates scores similar to ones produces by LoRA. Rows marked with superscript *†§ symbols correspond to values from prior works (Ben-Zaken et al., 2022; Houlsby et al., 2019; Rücklé et al., 2021) respectively. Value in parenthesis is a standard error rounded up to a single significant digit. The best top-2 values in a column are highlighted with bold. Higher is better.

| MODEL | METHOD | PARAM $10^3$ | RANK | CoLA | MNLI | MRPC | QNLI | QQP | RTE | SST2 | STS-B | Σ |
|---|---|---|---|---|---|---|---|---|---|---|---|---|
| RoBERTa$_{\text{base}}$ | FT | 125M | — | 61(1) | **87.6** | **89.3**(9) | 92.6(1) | **91.9** | **79**(2) | 94.1(1) | 90.4(2) | **86.4** |
| | BITFIT* | 113 | — | **62**(1) | 84.8(1) | **92.0**(4) | 91.3(2) | 84.5(2) | **78**(2) | 93.7(1) | 90.8(3) | 84.6 |
| | ADAPTER (RÜCKLÉ)§ | 300 | — | 60.8(4) | 87.1 | 89(1) | **93.1**(1) | 90.2 | 72(3) | **94.2**(1) | 89.7(3) | 84.4 |
| | | 900 | — | **62.6**(9) | **87.3**(1) | 88.4(1) | **93.0**(2) | **90.6** | 76(2) | **94.7**(3) | 90.3(1) | **85.4** |
| | LoRA | 295 | 8 | 61.1(6) | **87.3**(2) | 88(1) | 91.3(2) | 90.1(1) | 73(2) | **94.2**(2) | 90.7(2) | 81.5(8) |
| | LoTR | 74 | 32 | 60.5(?) | 85.2(6) | 85.9(4) | 90.0(1) | 87.4(1) | 66(4) | 93.0(4) | 88.8(4) | 77.2(9) |
| | | 100 | 40 | 58(2) | 85.2(2) | 88(1) | 92.5(3) | 87.6(0) | 53(14) | 93.8(7) | 89.8(5) | 78.2(5) |
| | | 276 | 80 | 61(2) | 84.6(1) | 89.0(0) | 92.1(5) | 86.8(0) | 71(3) | 93.4(1) | **90.9**(2) | 79.1(8) |
| | | 321 | 88 | 61.3(6) | 84.7(0) | 88.0(9) | 92.0(4) | 86.9(0) | 67(13) | 93.3(2) | **91.0**(1) | 79(2) |
| RoBERTa$_{\text{large}}$ | FT | 355M | — | **68** | 90.2 | **91** | 94.7 | **92.2** | **87** | **96.4** | **92.4** | **88.9** |
| | ADAPTER (HOULSBY)† | 6M | — | **67**(4) | 89.9(5) | **89**(3) | 94.7(2) | **92.1**(1) | 83(1) | 96.2(3) | 91(2) | **87.8** |
| | | 900 | — | 66(2) | **90.3**(3) | 88(2) | 94.7(2) | 91.5(1) | 73(3) | **96.3**(5) | 91.5(5) | 86.4 |
| | LoRA | 786 | 8 | 42(37) | **90.6**(2) | 75(12) | **94.8**(3) | 91.6(2) | 53(0) | 95.7(2) | **91.9**(4) | 69(8) |
| | LoTR | 328 | 64 | 61.3(9) | **90.3**(0) | **89.0**(5) | **94.8**(1) | 89.2(1) | **84**(2) | 95.9(1) | 91.6(1) | 83(9) |

with the number of dimensions rather than polynomially like $O(r^n)$ as HOSVD counterpart.

## 3. Experiments

We follow historically developed experimental methodology to benchmark our approach (Ben-Zaken et al., 2022; Dettmers et al., 2023; Guo et al., 2021; He et al., 2022; Houlsby et al., 2019; Hu et al., 2021). As a rule of thumb authors choose an encoder-only TRANSFORMER model like BERT (Devlin et al., 2019) or RoBERTa (Liu et al., 2019), then train it with different fine-tuning approaches on tasks from GLUE benchmark (Wang et al., 2018), and report results. Some authors (He et al., 2022) argue that experimenting with encoder-only models does not cover all aspects of fine-tuning of large language models and suggest to train encode-decoder and decoder-only models on language modeling tasks. In (Liu et al., 2022) authors consider few-shot learning. However, we stick to well-established practice and report fine-tuning of encoder-only models on GLUE (see Section 3.1). LoTR provides several knobs to adjust, among them are initialization scheme, multiplier $\alpha$, and a choice of weight matrices $W$. We address these factors in ablation studies in Section 3.2. Computation efficiency and practical aspects of LoTR are discussed in Section 3.3.

We use PYTORCH deep learning framework (Paszke et al., 2019) and pretrained models from HuggingFace's TRANSFORMERS library (Wolf et al., 2020) in all our experiments. We use ADAMW as a default optimizer (Loshchilov and Hutter, 2018). All experiments were carried out on several Nvidia GPUs of Ampere and Volta generations.

### 3.1. Encoder-only Models on GLUE

We follow (Hu et al., 2021) in evaluation of our approach on small and medium sized models and datasets. Namely, we train RoBERTa$_{\text{base}}$ and RoBERTa$_{\text{large}}$ models on datasets comprising GLUE benchmark (see Table 2).

We reproduce full fine-tuning experiments as reported in (Liu et al., 2019) for RoBERTa$_{\text{base}}$ with some small differences. First of all, we increase the number of training epoches from 10 to 20 since we noticed that convergence of LoTR and LoRA adapters is often slower and takes more than 10 epoches. Moreover, this number of training iterations is enough to observe the moment when a model starts to overfit (see Figure 3 and Figure 2). Next, we simplify experiments following (Hu et al., 2021) and fine-tune the original model checkpoint on MRPC, RTE, STS-B without auxiliary pretraining on MNLI.

We also reproduce LoRA on RoBERTa$_{\text{base}}$ and RoBERTa$_{\text{large}}$ models with hyperparameters reported in the original work (Hu et al., 2021). The only difference is that we limit the number of training epoches to 20 (as opposed to up 80 in LoRA paper) for all GLUE tasks. We find that the number of training epoches is a critical parameter for achieving the best scores and fair comparison. Since some LoRA experiments go far beyond boundary in 20 epoches, we carry out some additional runs varying learning rate in the small vicinity of the reported values and report the best scores on the validation set. Because of the limited computational budget we also reduce a number of repeated runs with different seeds from 5 to 3. This sometimes increases a standard error deviation (see LoRA scores on RoBERTa$_{\text{large}}$ CoLA in Table 2) which signifies of learning instability. We believe that the instability are caused by increased complexity in training over matrix manifolds.

In all experiments (except those in Section 3.2.3), we apply LoTR and LoRA adapters to $W_q$ and $W_v$ matrices. From general considerations, one might suggest that applying adapters to both query $W_q$ and key $W_k$ matrices does not make significant impact with respect to the number of parameters and score value. The reason is that matrices $W_q$ and $W_k$ are multiplied during computation of attention mask and their elements are not used at full expressivity. Prior work (Hu et al., 2021) confirms our reasoning empirically.

Evaluation of LoRA, LoTR, and other competitive approaches is presented in Table 2. It constitutes our main empirical result in this section. One can see that LoTR confidently demonstrates competitive performance. Comparison is carried out on two models RoBERTa$_{\text{base}}$ and RoBERTa$_{\text{large}}$. The second model is two times larger in the number of parameters and twice deeper than the first one (see Table 4). The difference of models in depth $L$ highlights that the joint representation of low-rank corrections as a tensor allows to achieve better parameter efficiency maintaining the same level of performance or even exceeding it. While LoTR is competitive with LoRA on RoBERTa$_{\text{base}}$, it confidently outperforms its counterparts on RoBERTa$_{\text{large}}$ model. Surprisingly, LoTR requires twice less parameters than LoRA to show the same score across GLUE tasks.

LoTR applied to RoBERTa$_{\text{base}}$ demonstrates worse performance metrics on downstream tasks but it is still impressive. Namely, rank 32 LoTR has only 74k trainable parameters (almost 5 times less than LoRA) and shows evaluation scores slightly lower than counterparts. In addition, we carry out experiments with LoTR adapters of ranks 80 and 88. These adapters pinch the LoRA adapter of rank 8 in the sense of the number of parameters (i.e. 276k < 295k < 321k). Thus, one can see that LoRA and LoTR with nearly the same number of parameters perform similarly on RoBERTa$_{\text{base}}$. Specifically, there are tasks on which LoTR is better than LoRA and vice versa (e.g. SST2 vs STS-B).

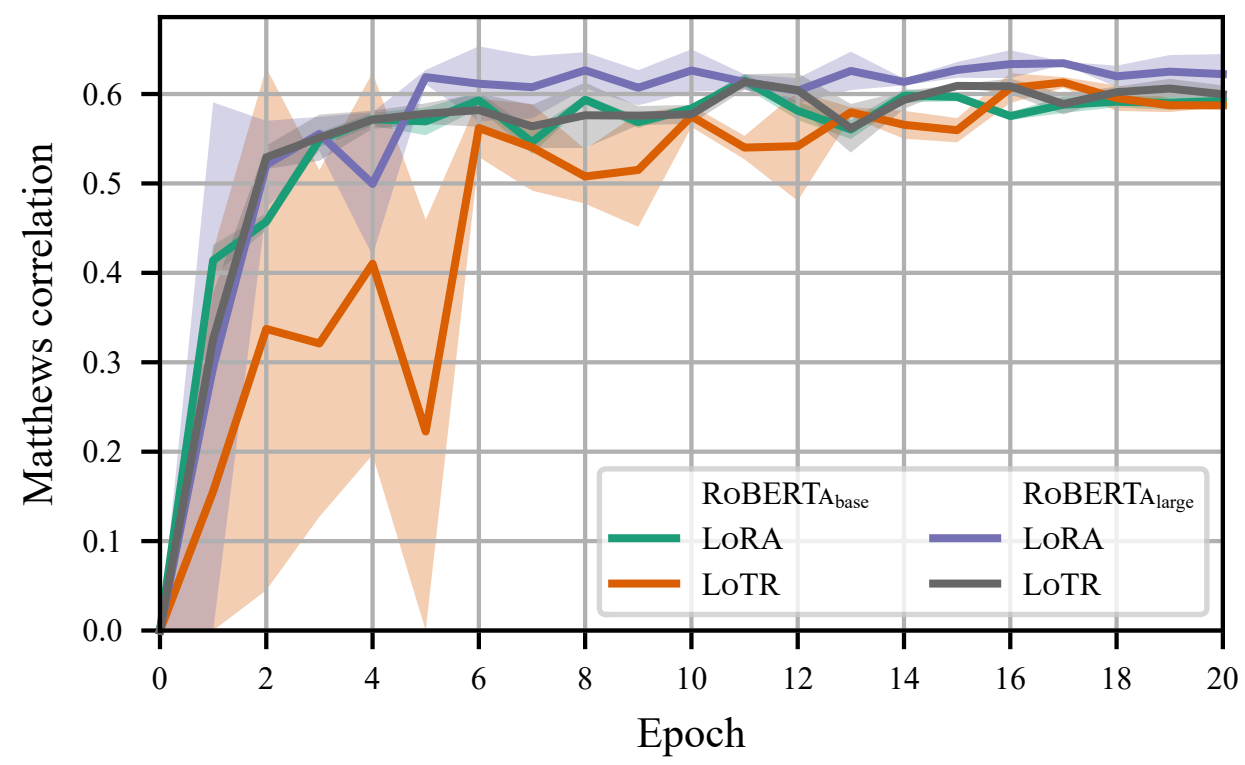

*Figure 2.* Evaluation metric for LoRA and LoTR adapters applied to RoBERTa models which are fine-tuned on CoLA. The best runs are presented.

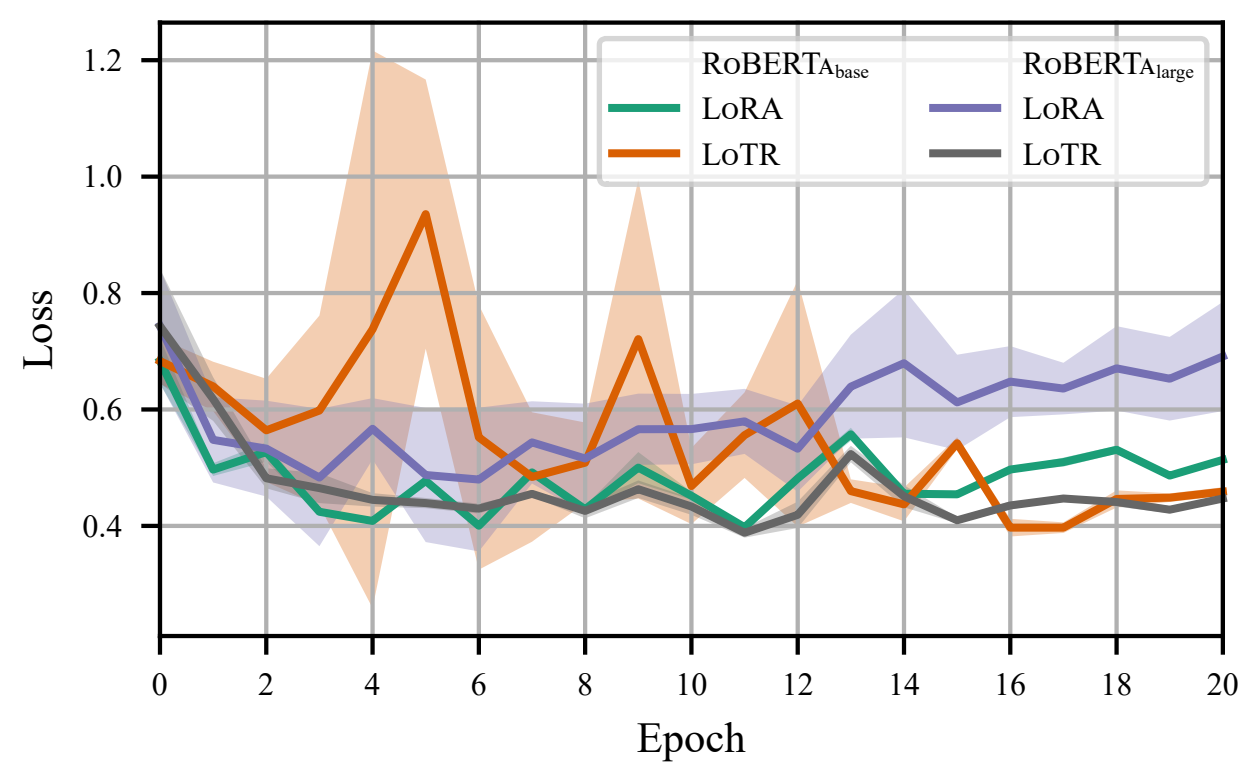

*Figure 3.* Training curve for LoRA and LoTR adapters applied to RoBERTa models which are fine-tuned on CoLA. The best runs are presented.

It is worth to note that there is a kind of scaling law in Table 2. Namely, evaluation scores increase as the number of trainable parameters grows. The most obvious way to trace the connection is the overall performance across tasks reported in the last column in Table 2.

## 3.2. Ablation Study

Our ablation study covers three major topics: (i) an impact of scaler $\alpha$ on convergence and evaluation metrics (see Section 3.2.1), (ii) an importance of weight initialization of LoTR adapter (see Section 3.2.2), and (iii) a choice of a parameter family for low-rank adaptation (see Section 3.2.3).

### 3.2.1. Scale $\alpha$

This section is devoted to empirical study on the scaler $\alpha$ and its impact on training and evaluation metrics. While the original work on LoRA (Hu et al., 2021) lacks a detailed discussion on the topic, it provides optimal value of $\alpha$ and general recommendations on tuning this hyperparameter.

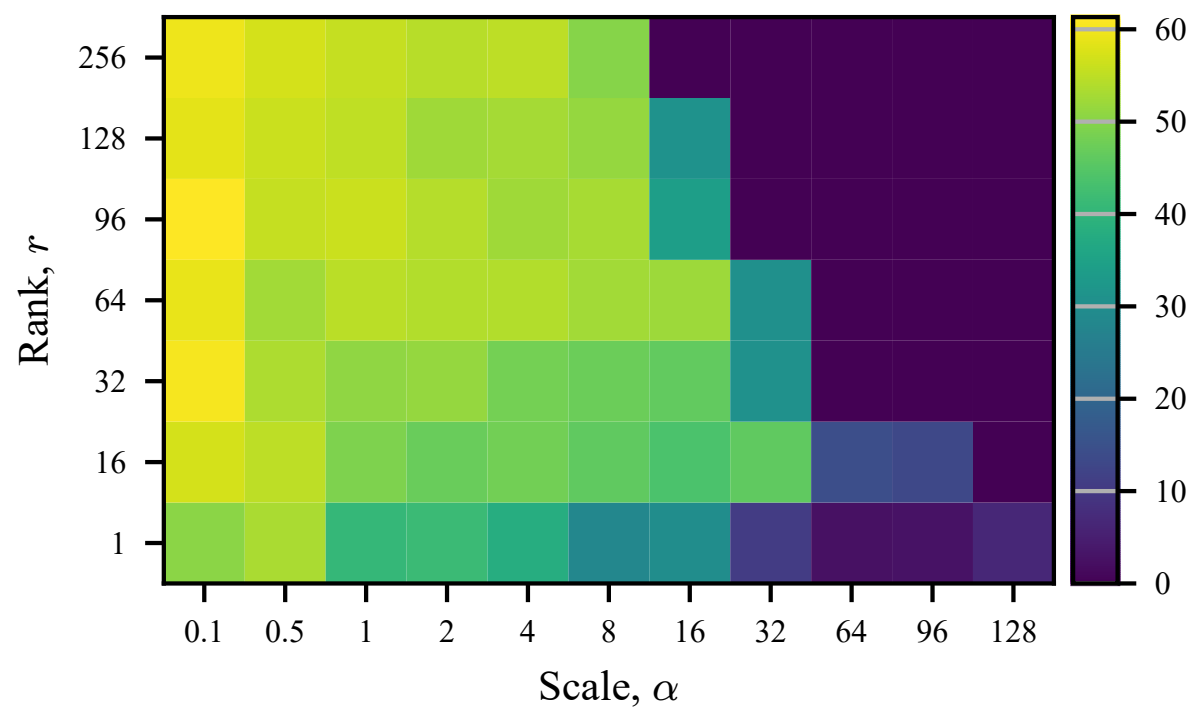

*Figure 4.* Influence of $\alpha$ scaler on target metric on CoLA downstream task with varying rank $r$ of LoTR. Higher is better. The best $\alpha$ is 0.1 on small tasks (e.g. CoLA or MRPC) and 1 on large tasks (e.g. MNLI or QQP).

While authors in (Hu et al., 2021) claim that a choice of $\alpha$ is insignificant, we found that in the case of LoTR $\alpha$ substantially influences convergence rate as well as the best achievable scores within a limited number of training iterations.

In order to clarify the role of $\alpha$ in fine-tuning of LoTR adapter, we carry out a series of experiments with RoBERTa$_{\text{base}}$ on the small datasets like CoLA and MRPC. We search the best pair of rank $r$ and scaler $\alpha$ over the grid of learning rates in the logarithmic scale (see Figure 4). We find that LoTR requires quite large learning rates of magnitude $10^{-3}$ and small $\alpha$ of order $10^{-1}$. Large values of $\alpha$ prevent convergence.

We verify our observations on bigger datasets like MNLI. However, instead of training a model for 20 epoches we train it for one epoch and use evaluation score as a proxy metric. Qualitative picture remains the same while optimal $\alpha$ drifts to bigger values of order $10^0$.

#### 3.2.2. Initialization

As it was discussed previously in Section 2.3.1, we consider several initializers for factor matrices $A$, $B$ and core tensor $G$ and study various initializer pairs empirically. Namely, we consider zero, normal, and Tucker2 initialization for factor matrices and zero, identity matrices (neutral), and Tucker2 initialization for core tensor. Although, exactly one of the terms $A$, $B$, or $G$ should be set to zero, we consider all 9 combinations of initializers.

In this study, we apply LoTR with different initialization choices to fine-tuning RoBERTa$_{\text{base}}$ on MRPC. The choice of MRPC is motivated by its small size in combination with sensitive target accuracy metric which allows us to reason about the original hypothesis. Hyperparameter and search spaces are reported in Appendix D.1.

*Table 3.* Comparison of LoTR applied to different sets of weight matrices. In first the first case, a single LoTR is applied to all $\{W_q, W_v\}$. In the second case, two LoTR adapters are applied to separate sets $\{W_q\}$ and $\{W_v\}$.

| Families | Num. Params | Rank | CoLA |
|---|---|---|---|
| $\{W_q, W_v\}$ | 100k | 40 | 58(2) |
| $\{W_q\}$ and $\{W_v\}$ | 123k | 32 | 56(3) |
| $\{W_q\}$ and $\{W_v\}$ | 161k | 40 | 58(6) |
| $\{W_q, W_v\}$ | 276k | 80 | 61(2) |

We find that the initialization of factor matrices $A$ and $B$ from standard normal distribution $\mathcal{N}(0, 1)$ and zeroed $G$ gives the best accuracy score on the validation set, while the rests stop at mediocre score values and do not converge to good minima. Surprisingly, the initializations with zeroed factor and without it demonstrate similar scores.

#### 3.2.3. Choice of Weight Family $\delta W$

In this section, we study how a choice of a set of corrections $\delta W$ and a number of adapter impacts performance on downstream tasks. Namely, we consider two scenarios. In the first one, LoTR adapter is applied to the whole weight matrices $\delta W_1 = \{W_q, W_v\}$ (i.e. there is only one adapter). In the second scenario, we apply two LoTR adapters to a set $\delta W_{2q} = \{W_q\}$ and a set $\delta W_{2v} = \{W_v\}$ separately. In other words, the total number of adapted matrices is the same but the number of adapters is different. Subsequently, the number of trainable weights and adapter ranks differ as well. The number of updated parameters in the first scenario is $2Lr^2 + 2dr$, while it is $2Lr^2 + 4dr$ in the second one. The latter scenario additionally introduces $2dr$ trainable weights more if the number of adapted matrices is the same.

We carry out several experiments in our common setup with different number of trainable parameters and adapter ranks applied to RoBERTa$_{\text{base}}$ on CoLA task (see Table 3). It turns out that rank of LoTR adapter influences performance on downstream tasks stronger than the number of adjustable parameters. Specifically, two adapters with rank 32 show worse metric than a single adapter of rank 40.

### 3.3. Computational Efficiency

A main concern regarding computational efficiency of LoTR is related to a core tensor $G$ of shape $r \times L \times r$, because each layer contributes a quadratic term $O(N_{\text{batch}} r_{\text{lotr}}^2)$ to time complexity. In comparison, LoRA is only linear in $r_{\text{lora}}$ with time complexity $O(N_{\text{batch}} d r_{\text{lora}})$. This means that LoTR requires more computational time than LoRA if $r_{\text{lotr}} > \sqrt{d r_{\text{lora}}}$ (see Section 2.3). Exact value of the threshold rank $r_{\text{threshold}}$ depends on hardware and software implementation. In fact, computational time and runtime memory usage are extremely sensitive and careful implementation is strongly desirable. Specifically, an ordering of matrix operations in forward and backward passes can significantly speed up a single training iteration as well as reduce memory usage (Cherniuk et al., 2023).

However, we undertake benchmarking of our naïve implementation to have a complete picture of the proposed method. We measure memory usage, throughput, and inference latency for LoRA and LoTR applied to RoBERTa models on MNLI dataset. At training time, LoTR consumes 1-2% more memory and slower than LoRA by 2-3%. At inference time, LoTR demonstrates slightly higher latency by ~0.5% than LoRA on Ampere generation GPU while the picture is reversed on older Volta GPUs (see Appendix C).

## 4. Related Works

The notion of parameter-efficient fine-tuning appeared in a series of pioneer works (Rebuffi et al., 2017; 2018; Rosenfeld and Tsotsos, 2020) which address the problem of transfer learning in computer vision domain. In these works authors developed an idea of an adapter, a small learnable auxiliary module. Specifically, the adapter they use is a small convolution layer plugged in parallel to original convolution layer.

The adapter idea was extended to natural language processing domain in (Houlsby et al., 2019) and subsequently has been studied in various works (Dettmers et al., 2023; He et al., 2022; Hu et al., 2021; Pfeiffer et al., 2021; Rücklé et al., 2021; Zhang et al., 2023). In (Houlsby et al., 2019) an adapter, additional parameters that are represented as a bottleneck residual block, is appended at the end of each Transformer-subblock. Adapter's residual block consists of two fully-connected layers with intermediate feature dimensions $m < d$.

A line of works (Dettmers et al., 2023; Hu et al., 2021; Lialin et al., 2023; Zhang et al., 2023) study low-rank adapters. (Hu et al., 2021) pioneered the area with LoRA, a simple adapter structure which is plugged in parallel and consists of multiplication of two matrices of incomplete rank. (Zhang et al., 2023) suggested an adaptive variant of LoRA, called AdaLoRA (previously known as MARVEL). It uses SVD for adapter representation and rank truncation operation. Practical aspects of a low-rank adapter-based approach are studied in (Cherniuk et al., 2023; Dettmers et al., 2023). The first work presented a scheme of training a low-rank adapter with original model weights in quantized arithmetics. The second one observed that a careful choice of matrix multiplication order in low-rank adapters significantly reduce compute and memory usage.

There is an alternative to adapters research direction (Ben-Zaken et al., 2022; Guo et al., 2021; Lester et al., 2021; Li and Liang, 2021; 2021; Liu et al., 2022; 2023). Paper (Ben-Zaken et al., 2022) reported that even small number of parameters in bias vectors are sufficient to get competitive performance. DiffPrunning introduced in (Guo et al., 2021) imposes a $L_0$-regularization constraint with differential relaxati on a vector of auxiliary parameters $\delta_\tau$ to reparametrize original parameters. (Liu et al., 2022) proposed to scale an intermediate activations in attention and linear layers. A connection between prompt fine-tuning methods (Lester et al., 2021; Li and Liang, 2021; Liu et al., 2023) and adapter methods are revealed in (He et al., 2022).

## 5. Comments and Discussions

We introduce LoTR, a novel parameter-efficient method for model alignment and domain adaptation. It expoits a low-rank structure of a tensor comprised of low-rank corrections to a weight matrices in attention module in Transformer-block. The closest counterparts to LoTR are LoRA and AdaLoRA which focus primarly on low-rank matrices rather than tensors. LoTR demonstrates competitive to LoRA performance in fine-tuning on downstream tasks of GLUE benchmark with comparable number of parameters. However, the main advantage of LoTR is that it is a priori tensorized. This means that we can achieve better parameter efficiency as well as naturally represent more complex or higher-order weight structures for fine-tuning.

Low-rank tensor representation further raises a question on a scaling law of downstream scores with respect to a number of trainable parameters. We find in our experiments that target metrics heavily depend on the total number of adjustable parameters. An in-depth study of parameter-efficient fine-tuning methods across models of different sizes might shed light on the subject and we leave it as a promising future work. Other prospective line of work is a development of rank adaptive parameter-efficient fine-tuning method. Although AdaLoRA is an examplar of matrix rank adaptive approach, direct generalization of AdaLoRA is not straight forward due to the curse of dimensionaly thus this issues reuiqres special techniques.

While we focus on the study of a low-rank tensor adaptation in general, practitioners might be interested in computational efficiency, engineering properties, and LoTR applicability in a general setup. Namely, efficient inference requires some engineering effort in implementation of optimized kernel. We believe that LoTR is fully compatible with quantization techniques like QLoRA and there is no obstacle which prevents LoTR to be applied across various models and datasets not limited to natural language domain.

## Acknowledgements

The work was supported by the Center in the field of Artificial Intelligence in the direction of optimizing management decisions to reduce the carbon footprint on the basis of the Skolkovo Institute of Science and Technology under Contract No. 70-2021-00145/10841 dated 02.11.2021 (items 3.3.2 and 3.3.4), Contract No. 10825/3978620 dated 26.08.2021 with Additional Agreement No. 3 dated 22.06.2023 (Application No 1).

## Impact Statement

This paper presents work whose goal is to advance the field of Machine Learning. There are many potential societal consequences of our work, none which we feel must be specifically highlighted here.

## A. Optimization over Matrix Manifold of Adapters

Consider a linear layer with weight kernel $W$ with low-rank correction $\delta W = USV^\top$. Vanilla optimization method like AdamW struggles to perform gradient optimization over parameters $U, S, V$ of low-rank correction $\delta W$. In order to mitigate this issue, we proposed gradient evaluation for correction $\delta W$ with the subsequent projection on low-dimensional manifold. Such modification of training algorithm incorporates an information about topological structure of a parameter space which consists of factorized matrices.

To be more specific, we calculate gradients

$$G_k = \frac{\partial L}{\partial \delta W^k} \tag{7}$$

in forward-backward passes for each $k$-th layer in a set as usual. Then we do retraction step which project gradient $G_k$ onto low-dimensional manifold of low-rank correction for all layers $l \in \{1, 2, ..., L\}$ at once. Retraction procedure can be formulated as the following optimization problem

$$\min_{U,V,\{S_k\}} \sum_{k=1}^{L} \|G_k - US_kV^\top\|^2. \tag{8}$$

and solved with ALS method.

We found that a Riemanian optimization over matrix manifold of LoRA parameters results in faster convergence rate. However, overfitting happens much earlier and target metric appears to be less by 1-2%%. We observe the similar picture in case of LoTR as well.

## B. Threshold LoTR Rank

In this section we consider the issue about how to express LoTR rank $r$ through parameters of other parameter-efficent fine-tuning method when the total number of training parameters are the same. We follow notation used in the main body: $L$ or $N_{\text{depth}}$ are the depth of Transformer, $d$ is an embedding size, ratio $\kappa = d/L$ is an characteristic (or effective?) size of a model, and rank $r$ is a rank of LoTR adapter.

It worth to note that all parameters are strictly positive. Specifically, $\kappa > 0$ and its typical values has an order of magnitude of $L$ (see Table 4).

### B.1 LoRA

According to Table 1, if the number of parameters $N_{\text{lotr}} = N_{\text{lora}}$ then

$$4Lr^2 + 8dr = 8Ldr_{\text{lora}}. \tag{9}$$

The solution of the quadratic equation is

$$\begin{aligned} r &= -\frac{d}{L} + \sqrt{\left(\frac{d}{L}\right)^2 + 2dr_{\text{lora}}} = \frac{d}{L}\left(\sqrt{1 + \frac{2L^2 r_{\text{lora}}}{d}} - 1\right) \\ &= \kappa\left(\sqrt{1 + \frac{2Lr_{\text{lora}}}{\kappa}} - 1\right) \approx \kappa\sqrt{\frac{2Lr_{\text{lora}}}{\kappa}} = \sqrt{2dr_{\text{lora}}}, \end{aligned} \tag{10}$$

where $\kappa = d/L > 1$ can be interpreted as a characteristic size of a model as a ratio of embedding size and depth of a Transformer (number of Transformer-blocks).

### B.2 Adapter (Houlsby)

According to Table 1, if the number of parameters $N_{\text{lotr}} = N_{\text{adapter}}$ then

$$4Lr^2 + 8dr = 2L(2dm + d + m), \tag{11}$$

*Table 4.* Typical values of main parameters which define models of common architectures. Parameter $\kappa$ can be viewed an order parameter or a compression rate of $\delta W$. It highlights a contribution of hidden dimension to the total number of trainable parameters.

| Model | Depth | Emb. Size | $\kappa$ |
|---|---|---|---|
| RoBERTa$_{\text{base}}$ | 12 | 768 | 64 |
| RoBERTa$_{\text{large}}$ | 24 | 1024 | 43 |
| LLaMA-7B | 80 | 8192 | 102 |

*Table 5.* Memory usage and throughput of LoRA and LoTR applied to RoBERTa models on MNLI dataset in our common setup. The number of parameters approximately the same in case of RoBERTa$_{\text{base}}$, LoRA has twice LoTR parameters in case of RoBERTa$_{\text{large}}$.

| Model | Method | Memory, MiB | Throughput, s$^{-1}$ |
|---|---|---|---|
| RoBERTa$_{\text{base}}$ | LoRA | 3790 | 10.17 |
| | LoTR | 3854 | 9.86 |
| RoBERTa$_{\text{large}}$ | LoRA | 9374 | 3.68 |
| | LoTR | 9458 | 3.62 |

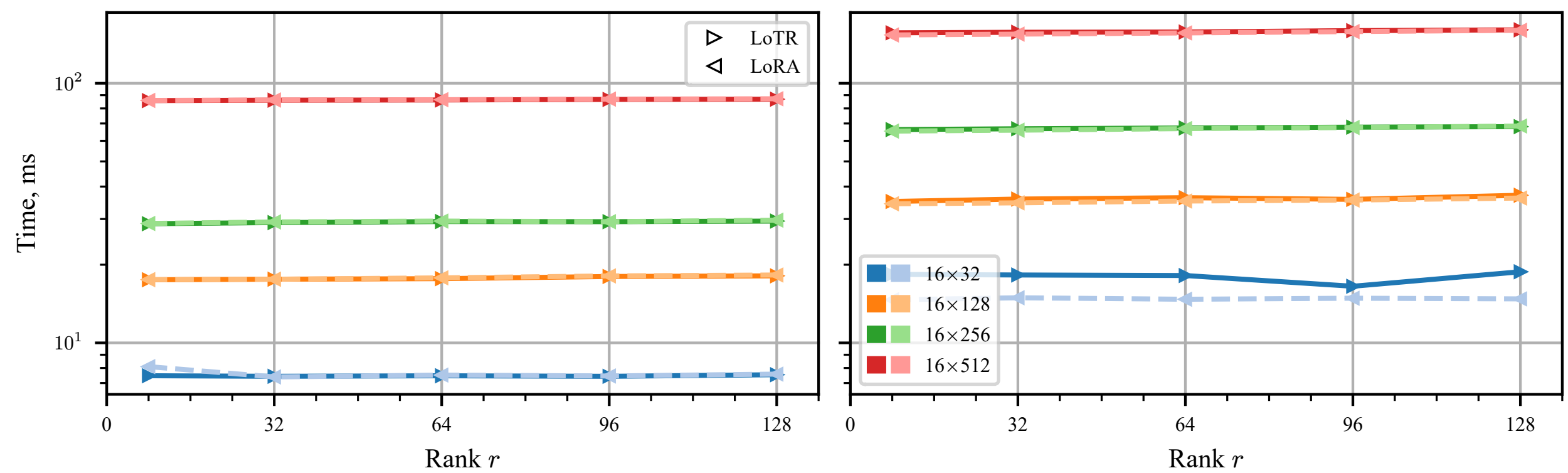

*Figure 5.* Comparison of inference time of LoRA and LoTR applied to RoBERTa$_{\text{base}}$. Batch is sampled from QQP dataset. It has fixed size and varying sequence length. Experiment is carried out on GPUus of Ampere (left) and Volta (right) generations. Trend lines overlap since inference times for both methods do not show a noticable different.

where $m$ is a embedding size in bottleneck. The soluition of the quadratic equation is

$$\begin{aligned} r &= -\frac{d}{L} + \sqrt{\left(\frac{d}{L}\right)^2 + dm + \frac{d+m}{2}} = \frac{d}{L}\left(\sqrt{1 + \left(\frac{L}{2}\right)^2 \left(dm + \frac{d+m}{2}\right)} - 1\right) \\ &= \kappa\left(\sqrt{1 + \left(dm + \frac{d+m}{2}\right)\kappa^{-2}} - 1\right) = \kappa\left(\sqrt{1 + \frac{Lm}{2\kappa}\left(2 + \frac{1}{d} + \frac{1}{m}\right)} - 1\right) \\ &\approx \kappa\sqrt{\frac{Lm}{\kappa}} = \sqrt{dm}, \end{aligned} \tag{12}$$

where $\kappa = d/L > 1$ is a characteristic size of a Transformer model. Curiously, $dm$ and $(d+m)/2$ can be interpreted as an area of a rectangle $dm$ and an area of a trapezoid of unity height with bases $d$ and $m$.

## C. Computational Efficiency

Table 5 presents runtime metrics of LoRA and LoTR adapters applied to RoBERTa$_{\text{base}}$ and RoBERTa$_{\text{large}}$ models on MNLI dataset. Figure 5 shows how inference time changes with adapter rank for different batches. Heatmap Figure 6 depicts a relative inference latency of LoTR with respect to LoRA.

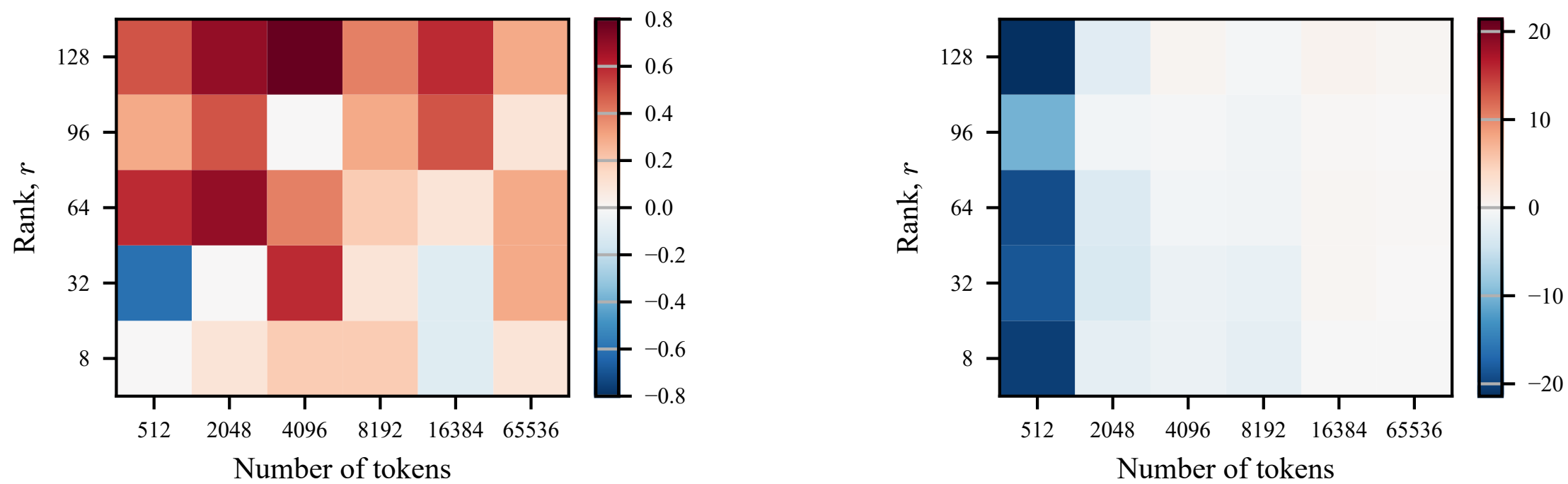

*Figure 6.* Relative inference latency difference between LoRA and LoTR, i.e. $(t_{\text{LoTR}} - t_{\text{LoRA}})/t_{\text{LoRA}}$. We use GPUs of Ampere (left) and Volta (right) generations. LoRA has longer lattency than LoTR on older GPUs. In general inference latency differs in a fraction of one percent.

## D. Hyperparameters

In this section we summarized hyperparameters used in experiments and search spaces for finding the best performing hyperparameters and the best scores.

### D.1 Initialization

We use our common fine-tuning pipeline and train a RoBERTa$_{\text{base}}$ model for 20 epoches on MRPC with linear warm up for 6% of steps and linear decay with standard batch size and learning rate (Liu et al., 2019). We pick rank 80 LoTR since we carry out the most of our experiments in the vicinity of this rank and total number of parameters is lower than corresponding LoRA adapter. All experiments are repeated tree times with different seed and then evaluation scores are averaged over runs. We extract the best (maximal) score across runs with different scalers $\alpha$ and learning rates. Scaler $\alpha$ and learning rate search spaces are chosen in a way that the most of best performing pairs of $\alpha$ and learning rates are in this square (Cartesian product of search spaces). In snippet below we use following notation: `trivial` means zeros initialization, `neutral` is identity matrices $G_s$, and `svd` corresponds to initialization from Tucker2 decomposition of frozen weights $\mathcal{W}$.

```
SEARCH_SPACE = {
    'task': ['mrpc'],
    'rank': [80],
    'scaler': [0.5, 1.0, 2.0, 8.0],
    'init_core': ['trivial', 'neutral', 'svd'],
    'init_factors': ['trivial', 'normal', 'svd'],
    'lr': [1e-2, 5e-3, 1e-3, 5e-4],
    'seed': [42, 3705, 0x12c946425095e587],
}
```

## E. Auxiliary Figures

In this section we place all figures which have not been fitted in the main body of the paper.

### E.1 Ablation Study

#### E.1.1 Hyperparameter $\alpha$

As it was mention previously, we carry out our experiments in the same setup. Figure 7 shows how target metric on CoLA changes over range of $\alpha$. Figure 8 demonstrate the dependence of evaluation score on rank $r$.

#### E.1.2 Parameter Scaling

Figure 9 shows the threshold LoTR rank $r_{\text{lotr}}$ for different counterpart. If LoTR rank $r < r_{\text{lotr}}$ then it requires less floating-point operations.

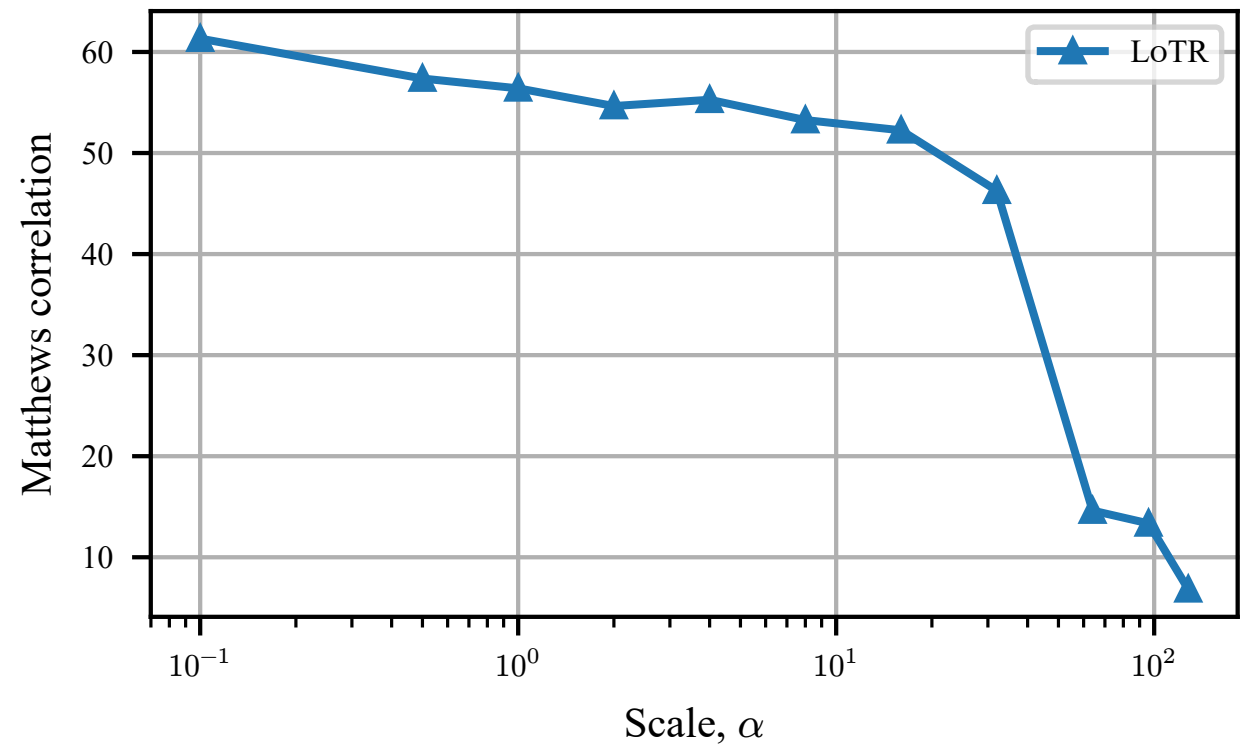

*Figure 9.* Influence of $\alpha$ scaler on target metric on downstream task (CoLA) with varying rank $r$ of LoRA and LoTR. There is a wide region on which suboptimal score values might be achieved.

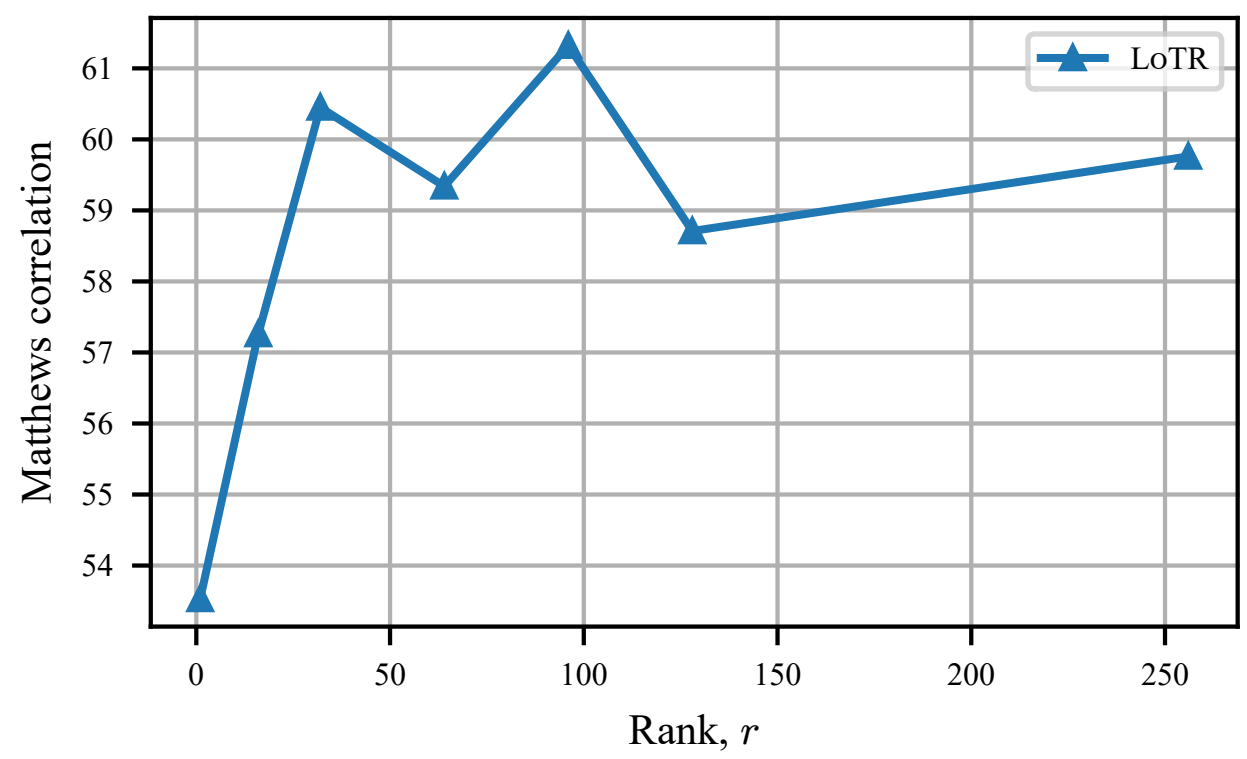

*Figure 9.* Dependency of a target metric on rank $r$ of LoTR applied to RoBERTa$_{\text{base}}$ model. Evaluation is done on CoLA task.

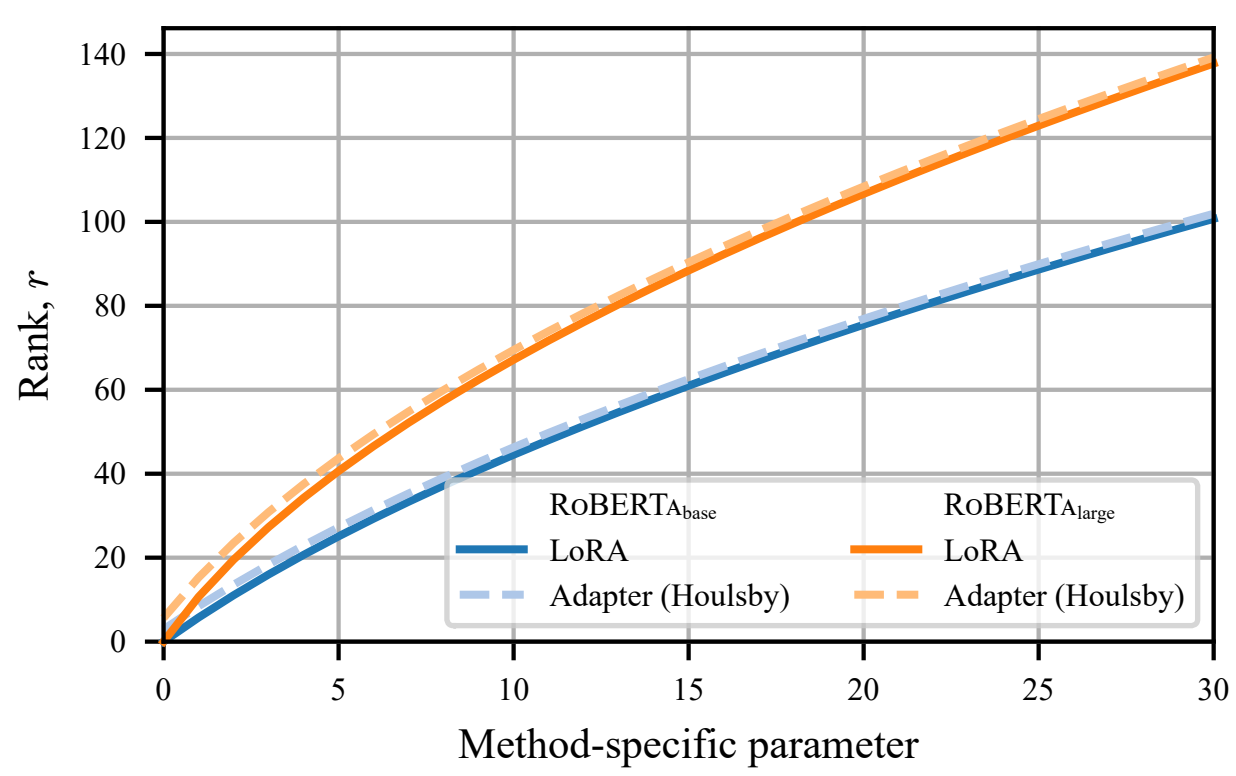

*Figure 9.* Relationship between LoTR rank $r$ and a parameter of concurrent method which gives the same number of updatable parameter (i.e. $N_{\text{lotr}} = N_{\text{method}}$).